\documentclass[letterpaper, 10 pt, conference]{ieeeconf}  

\IEEEoverridecommandlockouts                              

\usepackage[T1]{fontenc}
\usepackage[utf8]{inputenc}
\usepackage{graphicx}
\usepackage{amsmath}
\usepackage{amssymb}
\usepackage{tabularx}
\usepackage{booktabs} 
\usepackage{array}      
\usepackage{multirow}
\usepackage{subcaption}  
\usepackage{cite}
\usepackage{flushend}

\title{\LARGE \bf
Learning to Balance Motor Thermal Safety and Quadrupedal Locomotion Performance with Residual Policy
}

\author{
Yuhang Wan, Weixian Lin, Letian Qian, Yiqi Zou, Weiwei Wu, Shengwei Liao, Chuanlin Zhao, Xin Luo%
\thanks{
All authors are with the School of Mechanical Science and Engineering,
Huazhong University of Science and Technology,
Wuhan, 430074, China.}
\thanks{This work was supported in part by the National Natural Science Foundation of China (No. 52375014).}
}

\begin{document}

\maketitle
\thispagestyle{empty}
\pagestyle{empty}

\begin{abstract}

Motor thermal management is often overlooked in the context of electrically-actuated robots, particularly legged robots, but motor overheating is a key factor that limits long-duration locomotion especially under payload conditions. This paper integrates a whole-body thermal model of a quadruped robot into the reinforcement learning pipeline to update motor temperatures, and proposes a two-stage training framework for motor thermal management. In this framework, a nominal policy is first pre-trained as a locomotion baseline capable of traversing diverse terrains. A residual policy is then trained on top of the nominal policy to provide corrective actions based on the robot’s thermal state, ensuring high performance under low-temperature conditions and preventing motor overheating under high-temperature conditions. Simulation results demonstrate that the proposed policy achieves an effective balance between motor thermal safety and locomotion performance. Real-world experiments on a Unitree A1 quadruped robot further validate the approach: under a 3 kg payload, the robot achieves stable locomotion across multiple terrains for over 13 minutes, while the nominal policy alone leads to motor overheating in about 5 minutes.

\end{abstract}

\section{INTRODUCTION}

In the field of legged robots, reinforcement learning (RL) has become a widely adopted approach for high-performance locomotion, enabling electrically-actuated robots to be increasingly applied in heavy-load \cite{chang2025beyond,trivedi2025chance} and complex environment \cite{chen2024learning,qian2024leeps,wu2026toward} transportation tasks. However, existing RL-based policies usually assume that actuators can operate reliably at different thermal states. In practical tasks, 
during long-duration locomotion, sustained high current output leads to continuous heat accumulation in the motors. Once the temperature exceeds the safety threshold, low-level protection is triggered affecting the robot’s subsequent locomotion. In more severe cases, sustained overheating may lead to hardware damage. Therefore, actuator thermal constraints have increasingly become a critical bottleneck for sustained high-performance operation of robots.

Prior work has incorporated temperature-aware regularization into RL training to encourage thermally safe locomotion, primarily on flat terrains\cite{muller2025olaf,qian2026}. However, coupling locomotion performance and thermal safety objectives within a single policy introduces inherent limitations. Even within safe temperature ranges, the policy may still favor conservative locomotion style to avoid high current outputs\cite{qian2026}, thereby restricting adaptability in complex terrains. This phenomenon arises from the different styles of agile and thermally constrained locomotion, and a single neural network cannot simultaneously capture the action distributions of both styles\cite{liu2023improving,huang2025moe}. During training, temperature-related penalties are often assigned relatively high weights, which tend to drive the policy toward conservative behaviors, thereby preventing the robot from fully exploiting its performance potential under low-temperature conditions. Therefore, the central question addressed in this work is: how to balance thermal safety and locomotion performance, enabling high locomotion performance at low temperatures while preventing motor overheating at elevated temperatures.

In this work, we propose a two-stage RL framework with residual policy learning to address the action distribution shift\cite{li2025train,zhang2025motion} between high-performance locomotion and thermally constrained locomotion. First, a nominal policy is trained to ensure the robot’s fundamental terrain adaptability. Then, a residual policy is trained to modulate the nominal policy output with a thermal model of the quadruped robot integrated into our pipeline. We design reward functions related to temperatures and residual actions to encourage the residual policy to preserve the original dynamic performance at low temperatures, while progressively increasing corrective adjustments as joint temperatures rise. In this way, the proposed framework enables a balance between high-performance locomotion and thermal safety.

Our main contributions are summarized as follows:

\begin{itemize}

\item We propose a two-stage training framework including pre-training of a nominal locomotion policy to acquire essential terrain-adaptive locomotion abilities and a residual policy training stage for thermal management, during which a whole-body thermal model of the quadruped robot is incorporated into the training pipeline.

\item We design reward functions related to motor temperatures and residual actions, guiding the residual policy to adaptively adjust behavior at different thermal stages, achieving a balance between high-performance locomotion and thermal safety.

\item We conduct comprehensive evaluations in both simulation and hardware experiments, demonstrating the effectiveness of the proposed approach.

\end{itemize}

\section{RELATED WORKS}



\begin{figure*}[htbp]       
  \centering                
  \includegraphics[width=0.95\linewidth]{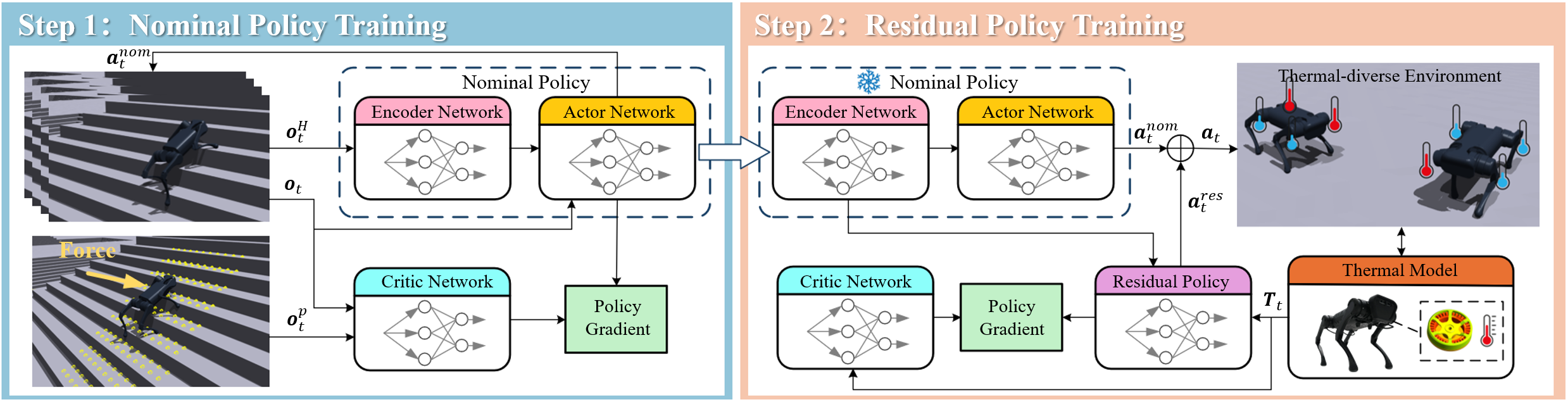}  
  \caption{Overview of the two-stage training framework: (1) pre-train a nominal policy as the baseline for locomotion; (2) integrate a whole-body thermal model of the quadruped robot into our pipeline and train a residual policy to modulate the nominal policy output.} 
  \label{fig:figure2}       
\end{figure*}

\subsection{Thermal Management of Robot Actuators}

Thermal management of robot actuators is typically addressed from hardware and software perspectives. On the hardware side, optimizing the robot’s heat dissipation system \cite{zhu2021design, akawung2020design} can increase the overall thermal capacity. However, such approaches do not alter the coupling between motor torque output and the current thermal state. When the robot operates continuously under high load, thermal management must still be handled at the software level. On the software side, thermal management is incorporated into the robot’s controller. Some studies employ motor thermal models to dynamically optimize joint torque outputs for temperature control \cite{kawaharazuka2020estimation}. With the development of RL, prior works have introduced temperature-related modulation into policy learning, incorporating thermal considerations through reward design \cite{muller2025olaf,qian2026}. In relatively simple scenarios, these approaches have been shown to improve the robot’s ability to operate for extended periods. However, the method in \cite{muller2025olaf} primarily regulates the temperature of the neck actuator. Meanwhile, the method in \cite{qian2026} applies temperature-related penalties within a unified policy, which encourages overheating avoidance but does not explicitly enable adaptive trade-offs between locomotion performance and thermal safety according to the robot’s thermal state.

\subsection{Residual Policy}

A residual policy enables local modulation of the nominal control output by adding a corrective term to the output. This structure typically models the fundamental capability and the additional regulation mechanism in a hierarchical manner, thereby enhancing system adaptability without altering the inherent style of the nominal policy. In prior work, the residual policy is typically employed to correct for dynamics model errors \cite{kim2025modular,he2025asap,lee2020learning}, improving the robustness and sim-to-real transfer performance on physical robots. In addition, some studies employ a high-frequency residual policy \cite{ma2025learning} to make precise adjustments to the nominal outputs. With the development of RL-based motion imitation methods, the residual policy has also been used to improve policy generalization and adaptability in action distributions outside the training dataset\cite{wang2026general}. For example, some studies first train a nominal policy with a fixed locomotion style based on reference data captured on flat terrain, and then introduce a residual policy to compensate it, enabling the robot to maintain its original locomotion characteristics while adapting to various complex terrains\cite{zhang2025motion}. Owing to its ability to reconcile significantly different action distributions, the residual policy holds considerable potential for modulating robot behavior based on long-term physical states, representing a promising direction for future research.

\section{METHODS}

\subsection{Framework Overview}

The proposed training framework for thermal-aware quadrupedal locomotion is shown in Fig. \ref{fig:figure2}. We construct a thermal model for the Unitree A1 and integrate it into our training pipeline to provide temperature feedback. The policy training proceeds in two stages. First, a nominal locomotion policy is trained to establish robust terrain-adaptive motion priors. Second, a thermal-aware residual policy is trained to modulate the nominal actions based on the robot’s thermal state, enabling adaptive regulation of actuator temperatures under high-temperature conditions while preserving locomotion performance at low temperatures.

\subsection{Nominal Policy Training}The training objective of the nominal policy $\pi_{\text{nom}}$ is to establish a locomotion baseline with reliable terrain-adaptive ability and agile performance, providing a fundamental motion capability for subsequent thermal-aware residual policy training. Considering that quadruped robot’s physical states under complex terrains are partially observable, relying only on instantaneous proprioceptive observation is insufficient to fully represent the system state. To address this issue, we adopt Asymmetric Actor–Critic framework \cite{pinto2017asymmetric} combined with an encoder to train the nominal policy effectively.

\textit{1) Nominal Policy Network:} The nominal policy consists of an encoder and an Actor. The encoder processes historical sequences of proprioceptive observation to extract latent environmental features, with its parameters optimized through Hybrid Internal Optimization (HIO) \cite{long2023hybrid}. The input to the actor consists of three components: the current proprioceptive observation $\mathbf{o}_t$, and the body velocity estimate $\hat{\boldsymbol{v}}_t$ and a latent vector $\hat{\mathbf{l}}_t$, both estimated by the encoder to capture the robot’s motion and environmental features. Specifically, the proprioceptive observation $\mathbf{o}_t$ is measured from joint encoders and IMU, and is defined as:

\begin{equation}
\mathbf{o}_t =
\big(
\boldsymbol{v}_t^{\text{cmd}},
\boldsymbol{\omega}_t,
\boldsymbol{g}_t,
\boldsymbol{\theta}_t,
\dot{\boldsymbol{\theta}}_t,
\boldsymbol{a}_{t-1}^{\text{nom}}
\big)
\label{equ1}
\end{equation}

\noindent where $\boldsymbol{v}_t^{\text{cmd}} \in \mathbb{R}^3$ is the commanded velocity in the robot's base frame. 
$\boldsymbol{\omega}_t \in \mathbb{R}^3$ and $\boldsymbol{g}_t \in \mathbb{R}^3$ denote the angular velocity and gravity vector in the robot's base frame, respectively. 
$\boldsymbol{\theta}_t \in \mathbb{R}^{12}$ and $\dot{\boldsymbol{\theta}}_t \in \mathbb{R}^{12}$ represent the joint positions and velocities, and 
$\boldsymbol{a}_{t-1}^{\text{nom}} \in \mathbb{R}^{12}$ is the previous actions output by the nominal policy. 
The encoder estimates the body linear velocity $\hat{\boldsymbol{v}}_t$ and a latent vector $\mathbf{l}_t \in \mathbb{R}^{16}$ from a historical sequence of six proprioceptive observation frames $\mathbf{o}_t^{H}$. The implementation details of HIO can be found in \cite{long2023hybrid}.

\textit{2) Critic Network:} To achieve more accurate state-value estimates, the critic receives not only the proprioceptive observation $\mathbf{o}_t$ but also additional privileged observations $\mathbf{o}_t^{p}$ during training. 
These privileged observations include the body velocity $\boldsymbol{v}_t \in \mathbb{R}^3$, the external forces applied to the body $\boldsymbol{F}_t \in \mathbb{R}^3$, and surrounding ground height $\boldsymbol{m}_t \in \mathbb{R}^{187}$.

\textit{3) Action Space:}
The action space of the nominal policy represents the joint position offsets of the quadruped robot relative to their default positions. 
The target joint positions are determined as
$\boldsymbol{\theta}_{\text{target}} =
\boldsymbol{\theta}_0 + \boldsymbol{a}_t^{\text{nom}}$, and converted into joint torques by PD controllers, where $\boldsymbol{\theta}_0$ denotes the default joint positions and $\boldsymbol{a}_t^{\text{nom}} \in \mathbb{R}^{12}$ is the actions output by the nominal policy.

\textit{4) Reward Function:} Following previous studies on blind locomotion of quadruped robots \cite{long2023hybrid,lee2020learning}, the reward function is summarized in Table~\ref{tab:table1}, which does not include any motor temperature-related terms and is designed to optimize the robot’s locomotion performance across different terrains and working conditions. The training details of the nominal policy is further described in Section \ref{Training-Details}.

\renewcommand{\arraystretch}{1.5}
\begin{table}[htbp]
\begin{center}
\captionsetup{
    labelsep=newline, 
    justification=centering, 
    singlelinecheck=false,
    font=small,              
    labelfont=normalfont,    
    textfont=sc              
  }
\caption{REWARDS FOR NOMINAL POLICY TRAINING}
\label{tab:table1}
\small 
\begin{tabularx}{\columnwidth}{
  >{\centering\arraybackslash}m{0.34\columnwidth}
  >{\centering\arraybackslash}m{0.45\columnwidth}
  >{\centering\arraybackslash}m{0.12\columnwidth}
}
\toprule
Reward Item & Equation & Weight \\
\midrule

Lin. velocity tracking &
$\exp\!\Big(-\frac{\|\boldsymbol{v}_{xy}^{\mathrm{cmd}} - \boldsymbol{v}_{xy}\|_2^2}{\sigma}\Big)$ & 1.0 \\

Ang. velocity tracking &
$\exp\!\Big(-\frac{(\omega_{\mathrm{yaw}}^{\mathrm{cmd}} - \omega_{\mathrm{yaw}})^2}{\sigma}\Big)$ & 0.5 \\

Lin. velocity (z) & $v_z^2$ & -2.0 \\

Ang. velocity (xy) & $\boldsymbol{\omega_{xy}}^2$ & -0.05 \\

Orientation & $\|\mathbf{g}\|_2^2$ & -0.2 \\

Joint accelerations & $\|\ddot{\boldsymbol{\theta}}\|_2^2$ & -2.5e-7 \\

Termination & / & -200 \\

Body height & $(h^{\mathrm{target}} - h)^2$ & -1.0 \\

Foot clearance &
$\sum_{i=0}^{3} ( p_z^{\mathrm{target}} - p_z^i )^2 \cdot v_{xy}^i$ & -0.01 \\

Action rate &
$\|\boldsymbol{a}_t^{\text{nom}} - \boldsymbol{a}_{t-1}^{\text{nom}}\|_2^2$ & -0.01 \\

Smoothness &
$\|\boldsymbol{a}_t^{\text{nom}} - 2\boldsymbol{a}_{t-1}^{\text{nom}} + \boldsymbol{a}_{t-2}^{\text{nom}}\|_2^2$ & -0.01 \\

\bottomrule
\end{tabularx}
\end{center}
\end{table}

\noindent where the scaling factor for velocity tracking $\sigma = 0.25$, the target base height $h^{\text{target}} = 0.38\text{m}$, and the target foot clearance $p_z^{\text{target}} = 0.2\text{m}$.

\subsection{Thermal-Aware Residual Policy Training}

After training the nominal policy $\pi_{\text{nom}}$ as a locomotion prior, we train a residual policy to enforce actuator thermal safety while preserving the locomotion performance of the nominal policy. During this stage, the nominal policy $\pi_{\text{nom}}$ learned in the previous phase is frozen and only outputs nominal actions $\boldsymbol{a}_t^{\text{nom}}$. The residual policy $\pi_{\text{res}}$ generates corrective actions $\boldsymbol{a}_t^{\text{res}}$, and the final joint commands are obtained by linearly combining two actions:

\begin{equation}
\boldsymbol{\theta}_{\text{target}} =
\boldsymbol{\theta}_0 + \boldsymbol{a}_t^{\text{nom}} + \boldsymbol{a}_t^{\text{res}}
\label{equ2}
\end{equation}

\begin{figure}[htbp]       
  \centering                
  \includegraphics[width=0.95\linewidth]{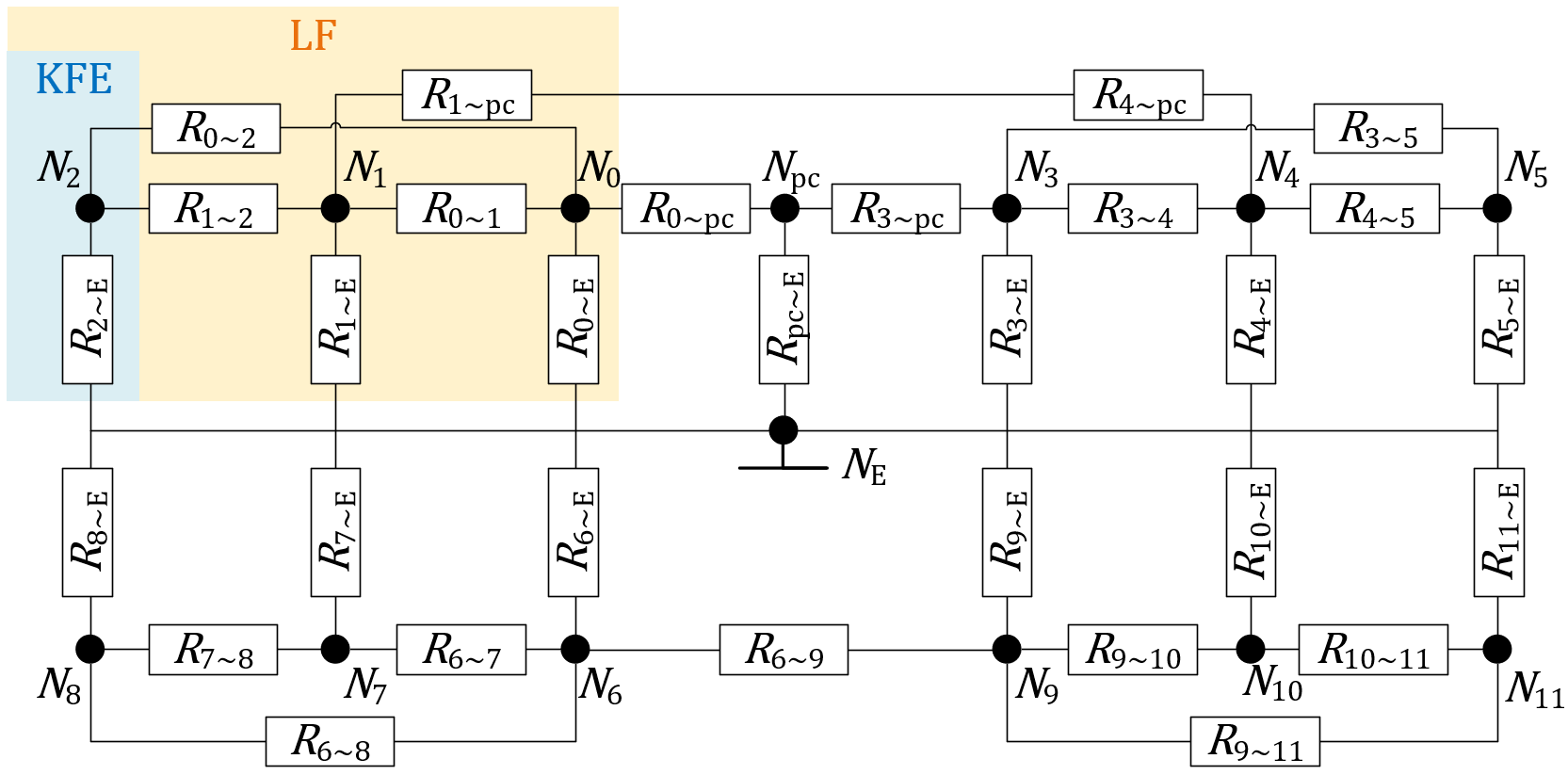}  
  \caption{Whole-body thermal model of the Unitree A1 robot.}          
  \label{fig:figure3}       
\end{figure}

\textit{1) Integration of Quadrupedal Thermal Model into the training Pipeline:} To address the lack of actuator thermal models in simulator for RL, we construct a whole-body thermal model for the Unitree A1 shown in Fig.~\ref{fig:figure3} according to \cite{LIN2025} and integrate it into our training pipeline, enabling real-time updates of motor temperatures for each agent.

$N_i$ represents thermal nodes, including 12 motors, the onboard computer, and the ambient environment. The thermal resistance $R_{i \sim j}$ between nodes i and j represents the equivalent resistance for heat transfer between two nodes. The heat generation $Q_{in,i}$ at each motor node primarily comes from Joule heating caused by the motor current, constant heat generated by the actuator driver, and mechanical friction heat generated by joint motion. Meanwhile, the model incorporates a forced convection correction mechanism based on the robot’s velocity to dynamically adjust the thermal resistance parameters.

\begin{equation}
C_i \, \dot{T}_i =
- \frac{T_i - T_E}{R_{i \sim E}(v_{xy})}
- \sum_{j \in \mathcal{M}(i)} \frac{T_i - T_j}{R_{i \sim j}}
+ Q_{\text{in},i}
\label{equ3}
\end{equation}

\noindent where $R_{i \sim E}(v_{xy})$ is the convection thermal resistance, $C_i$ is the thermal capacitance, and 
$\mathcal{M}(i)$ denotes the set of all neighboring thermal nodes connected to node $N_i$. 

To integrate the thermal dynamics into our pipeline, the continuous-time thermal model is discretized at the thermal update interval. Under a zero-order hold assumption on the heat-generation input $\boldsymbol{u}_t$, the RMS motor torques are used as equivalent inputs within each update interval, since motor torques vary faster than the thermal model updates. The resulting model is expressed in the following discrete-time MIMO state-space form:

\begin{equation}
\boldsymbol{T}_{t+1} = \mathbf{A}(v_{xy,t}) \, \boldsymbol{T}_t + \mathbf{B} \, \boldsymbol{u}_t
\label{equ4}
\end{equation}

\noindent where $\mathbf{A}(v_{xy,t}) \in \mathbb{R}^{14 \times 14}$ is the system matrix, capturing the thermal coupling between nodes, and $\mathbf{B} \in \mathbb{R}^{14 \times 14}$ is the input matrix describing the effect of motor inputs on temperature dynamics. The system input $\boldsymbol{u}_t$ corresponds to the heat generated by the motors, computed from their speeds and the RMS motor torques over the corresponding thermal update interval.

The thermal parameters are adopted from~\cite{LIN2025}, where they were identified by least-squares fitting using motor temperature data collected during natural cooling, prescribed joint-torque commands, and locomotion trials. The reported model achieves a mean squared error (MSE) below 2.5~($^\circ$C)$^2$ across different operating conditions, including locomotion with a 4 kg payload. In this work, we use this identified model for real-time temperature updates during RL training.

\textit{2) Residual Policy:} The residual policy $\pi_{\text{res}}$  serves as an independent compensation layer, modulating the nominal outputs. The observation of the residual policy is defined as:

\begin{equation}
\mathbf{o}_t^{\text{res}} =
\big(
\boldsymbol{v}_t^{\text{cmd}},
\boldsymbol{\omega}_t,
\boldsymbol{g}_t,
\boldsymbol{\theta}_t,
\dot{\boldsymbol{\theta}}_t,
\boldsymbol{T}_t,
\hat{\mathbf{l}}_t,
\boldsymbol{a}_{t-1}^{\text{res}}
\big)
\label{equ5}
\end{equation}

\noindent where $\boldsymbol{T}_t \in \mathbb{R}^{12}$ represents the real-time motor temperatures, and $\hat{\mathbf{l}}_t$ is the latent vector output by the encoder of the nominal policy, used to perceive the current terrain. In addition to proprioceptive observation, the Critic also receives privileged observations during training, including persistent external forces and surrounding ground height.

\textit{3) Reward Design:} To balance thermal safety and locomotion performance, we design a temperature-related reward function that guides the residual policy to intervene only when thermal risks arise. The reward function is composed as follows:

\begin{equation}
R_{\text{res}} = R_{\text{th}} + R_{\text{reg}} + R_{\text{nom}}
\label{equ6}
\end{equation}

\textbf{Thermal Safety Reward $\boldsymbol{R}_{\text{th}}$}. Due to the large thermal inertia of the actuators, temperature dynamics respond much more slowly than the robot’s mechanical dynamics, making it difficult to form a complete cooling feedback loop within a single episode. We design an exponentially weighted reward based on the temperature change rate $\dot{T}$, which encourages the policy to avoid further temperature rises of high-temperature motors and actively take actions to reduce motor temperatures. Importantly, the reward is determined by the temperature changes resulting from the robot’s behaviors, rather than the temperature values themselves.

\begin{equation}
R_{\text{th}} =
\omega_{\text{th}} \sum_{i=1}^{12} \dot{T}_i \,
\exp\Big( - \min \big( \sigma_{\text{th}} (T_{\max} - T_i), 0 \big) \Big)
\label{equ7}
\end{equation}

\noindent where $T_{\max} = 60^\circ\text{C}$ is the safety threshold, $\sigma_{\text{th}} = 0.35$ is the sensitivity coefficient, and $\omega_{\text{th}} = -1000$ is the weight of this reward term. At low temperatures, the penalty of this reward is almost zero, allowing the policy to ignore the motor temperature and focus on satisfying other reward terms. When the actuator temperature exceeds around $50^\circ\text{C}$, the weight of this reward term increases exponentially, forcing the policy to prioritize reducing the temperature of these actuators.

\textbf{Residual Action Regularization Reward $\boldsymbol{R}_{\text{reg}}$}. To encourage minimal modifications to the nominal policy, we introduce a penalty term based on the $\ell_2$ norm of the residual actions with a weight of $\omega_{\text{reg}} = -0.1$. This regularization guides the residual outputs to converge to zero especially at low temperatures, thereby fully preserving the original locomotion style of the nominal policy.

\begin{equation}
R_{\text{reg}} = \omega_{\text{reg}} \, \|\boldsymbol{a}_t^{\text{res}}\|_2^2
\label{equ8}
\end{equation}

\textbf{Nominal Policy Task Reward $\boldsymbol{R}_{\text{nom}}$}. We directly inherit the full reward function used during nominal policy training to ensure that the residual policy maintains the robot’s locomotion capability while modulating the nominal policy. Notably, during residual policy training, all action-related penalty terms in the original reward function are no longer applied to nominal actions $\boldsymbol{a}_t^{\text{nom}}$, but to residual actions $\boldsymbol{a}_t^{\text{res}}$.

\subsection{Training Details} \label{Training-Details}

We use the Isaac Gym simulator to build the training environment. During nominal policy training, 4096 parallel agents are used to train  a stable locomotion baseline. In the residual policy training stage, the number of parallel agents is increased to 16384 to cover a wide range of motor temperature samples. The nominal policy, residual policy, and full-body thermal model run synchronously at 50 Hz, while the low-level PD controller operates at 200 Hz with $K_p = 40.0$ and $K_d = 1.0$.

To improve policy robustness and explicitly generate conditions that lead to motor overheating, we introduce domain randomization of payloads and external disturbances during training. In each agent, the robot’s body center of mass is randomly shifted within $\pm 0.05~\text{m}$ along all three axes, and an additional mass of $[0,5]~\text{kg}$ is applied to the body. On top of this, a persistent external force with components randomly sampled from $[-30,30]~\text{N}$ is applied at the base. During residual policy training, motor initial temperatures are randomly initialized within $[T_{\max}-25^\circ\text{C}, T_{\max}+10^\circ\text{C}]$, and ambient temperatures are sampled within $[0, 35]~^\circ\text{C}$, ensuring that the policy is trained with abundant high-temperature samples.

During training, the robot’s initial terrain and commanded velocities are randomly sampled. The terrains include rough surfaces, slopes, stairs, and discrete obstacles, while the sampling range of linear and angular velocities start at $[-1, 1]~\text{m/s}$ and $[-2, 2]~\text{rad/s}$. We use a similar terrain curriculum as \cite{long2023hybrid,rudin2022learning}: the terrain difficulty is progressively increased as the robot moves sufficiently far from its starting position, and the $x$-direction velocity sampling range is gradually expanded to $[-2, 2]~\text{m/s}$ when the velocity tracking reward achieves a predefined threshold.

\section{EXPERIMENTS AND RESULTS}

The main advantage of our policy is its ability to maintain agile and terrain-adaptive locomotion at low temperatures while ensuring thermal safety at high temperatures. To validate its effectiveness, we conducted a series of simulation and physical experiments, comparing it with the following two policies. The simulation experiments were conducted in the Isaac Gym simulator, with temperature provided by the thermal model used in this work. The physical experiments were performed on the Unitree A1, with temperature obtained from the onboard motor temperature sensors.

\textbf{Nominal Locomotion Policy (NLP)}: This policy which refers to \cite{long2023hybrid} corresponds to the nominal policy component of our framework, serving as a baseline representing high-performance locomotion capable of traversing diverse terrains.

\textbf{Monolithic Thermal Policy (MTP)}: The MTP is trained using the same architecture and training procedure as the nominal policy, with proprioceptive observations extended to include all motor temperatures. Initially, it is trained for 6000 steps on complex terrains with nominal policy reward $R_\text{nom}$ to obtain a locomotion policy capable of handling diverse complex terrains. It is then fine-tuned with the thermal safety reward $R_\text{th}$ incorporated on nearly flat terrain with height variations of $\pm 2\,\text{cm}$ and a maximum slope of $5^\circ$ \cite{valsecchi2024accurate}.

\begin{figure}[!htbp]       
  \centering                
  \includegraphics[width=0.85\linewidth]{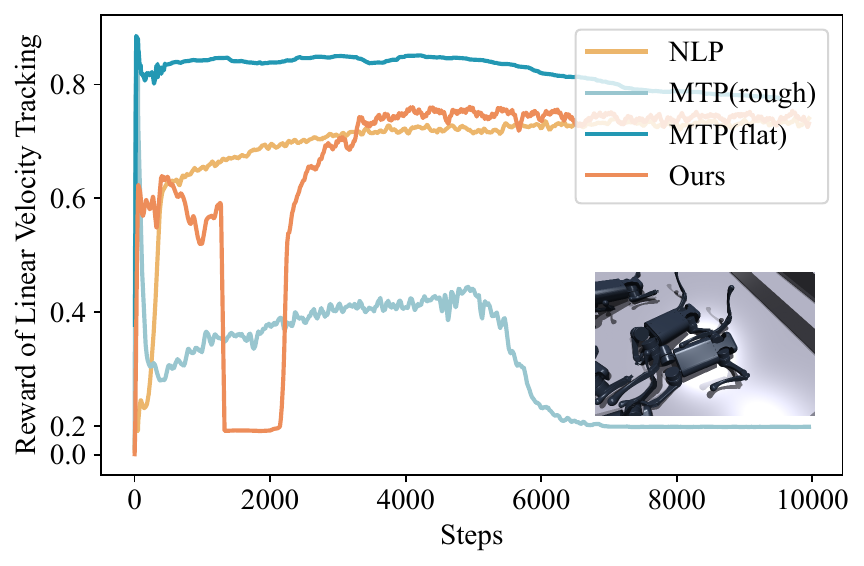}  
  \caption{The evolution of the linear velocity tracking reward during training. The image above the light-blue curve illustrates the deployment performance of the MTP after 10,000 steps of training on complex terrain.}          
  \label{fig:figure4}       
\end{figure}

The domain randomization settings and reward weights for the above policies were kept consistent with those used in our policy. Fig.~\ref{fig:figure4} shows that if the MTP is fine-tuned on complex terrains, the velocity tracking reward drops significantly, and the robot ultimately remains stationary. This drop is due to the substantial difference between the action distributions of terrain-adaptive locomotion and thermally constrained motion. To obtain a locomotion policy suitable for comparison, in this study the MTP was trained on nearly flat terrain. By contrast, our residual policy is able to compensate for this distribution mismatch, gradually restoring the tracking performance.

\subsection{Evaluation with Simulation}

\textit{1) Long-horizon Locomotion on Flat Terrain:} We conducted long-horizon locomotion experiments in the Isaac Gym simulator with 4096 robots running in parallel.For each robot, the payload, persistent external forces, and ambient temperature were sampled from the same domain randomization ranges as those used during training. The velocity commands were sampled within $[-2, 2]~\text{m/s}$ in the x-direction, $[-1, 1]~\text{m/s}$ in the y-direction, and $[-2, 2]~\text{rad/s}$ in yaw. The robots continuously walked for 800 s on nearly flat terrain with command updates every 30 s, and all motors were initialized at $20^\circ\text{C}$.

\begin{figure}[!htbp]       
  \centering                
  \includegraphics[width=0.90\linewidth]{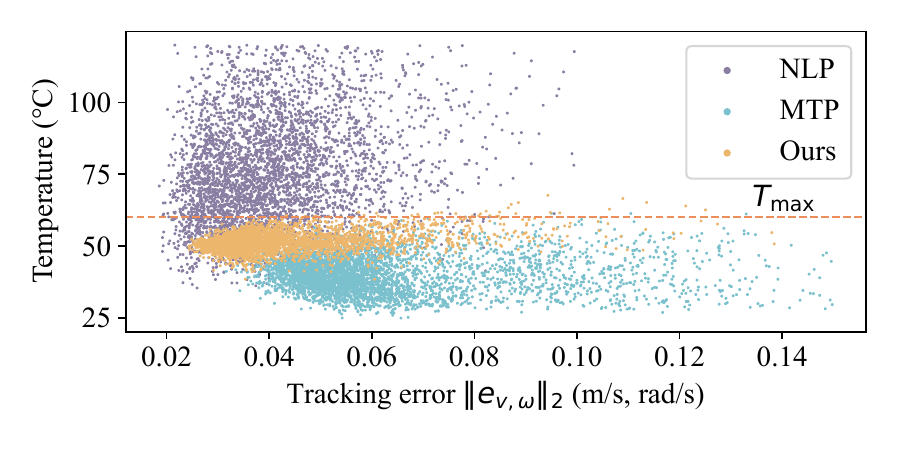}  
  \caption{Distribution of the robot’s performance over 800 s of locomotion under the three policies.}          
  \label{fig:figure5}       
\end{figure}

We recorded each robot’s velocity tracking error and the maximum motor temperature during the locomotion, as shown in Fig.~\ref{fig:figure5}. The results indicate that although NLP maintained very low tracking errors, many samples exceeded the safety temperature threshold, posing a significant risk of motor overheating. In contrast, MTP effectively reduced the overall temperature distribution, but its tracking errors showed considerable variation within the range $[0.03, 0.14]$. In contrast, our policy maintained tracking accuracy close to NLP while keeping the peak motor temperatures for most robots near but below the safety threshold, achieving near-optimal locomotion performance under thermal constraints.

\begin{figure}[t]       
  \centering                
  \includegraphics[width=1.0\linewidth]{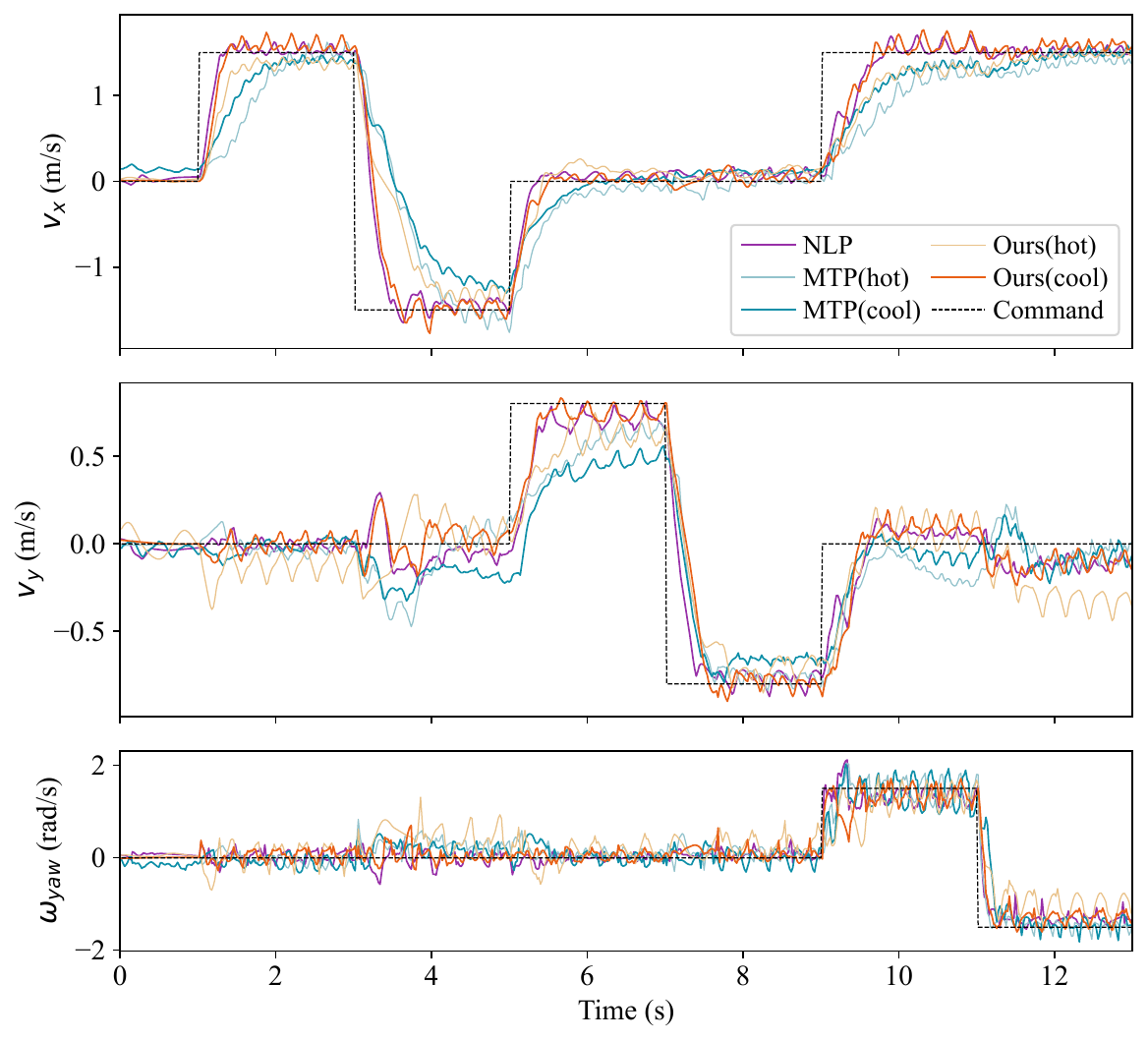}  
  \caption{Velocity curves of the robot under the three policies with commands for the x, y, and yaw directions applied simultaneously in high-temperature and low-temperature conditions.}          
  \label{fig:figure6}       
\end{figure}

\textit{2) Performance under Step Velocity Commands:} We further evaluated the performance of the robots under step velocity commands for different policies, with all motors initialized at low temperature ($30^\circ\text{C}$) and high temperature ($58^\circ\text{C}$). As shown in Fig.~\ref{fig:figure6}, MTP exhibited slow responses to velocity steps at both temperatures especially in the x and y directions, suggesting that it adopts a globally conservative walking style that fails to fully exploit the robot’s hardware capabilities under low-temperature conditions. In contrast, under the low-temperature condition, our policy and the NLP exhibited nearly identical response speeds along the x, y, and yaw axes. This demonstrates that the residual term only makes minor adjustments within the safe thermal range. At high temperatures, our policy showed reduced step response performance in the x and y directions, accompanied by relatively large fluctuations along the y-axis and in yaw. This behavior results from the residual policy, which modulates control outputs to regulate actuator temperatures. As shown in the supplementary material, this thermally adaptive way did not disturb the robot’s stability.

\textit{3) Terrain Adaptation:} We conducted locomotion experiments in two types of terrain: stairs (height 0.1 m and width 0.3 m) and slopes ($20^\circ$ slope), each with a horizontal length of 6 m. For each terrain, we performed 300 trials to evaluate the robots’ performance under different strategies, where they carried a 3 kg payload and moved at a commanded velocity of 1 m/s in the x-direction with initial motor temperatures set to $30^\circ\text{C}$, $50^\circ\text{C}$, and $58^\circ\text{C}$ respectively. A trial was classified as successful if the robot successfully traversed the terrain, overheated if any motor exceeded the threshold temperature during locomotion, drifting if the robot deviated from the planned path, failed if the robot’s torso made contact with the environment within its patch, and stuck if the robot had no effective movement within 30 s.

\begin{figure}[t]
  \centering
  \begin{subfigure}[b]{1.0\linewidth}
    \centering
    \includegraphics[width=\linewidth]{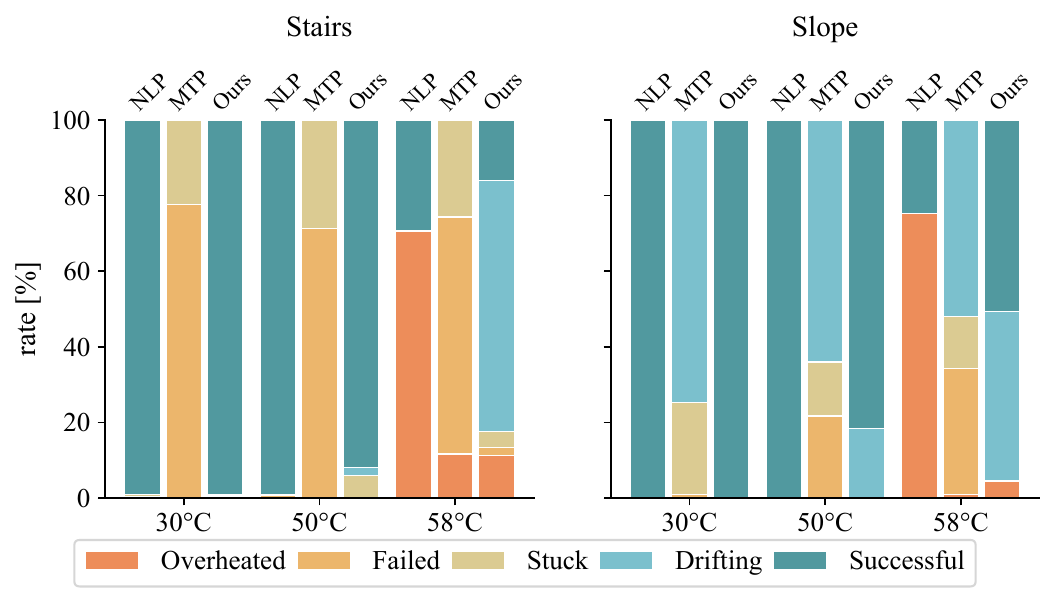}
    \caption{Distribution of the experimental outcomes.} 
    \label{fig:fig7a}
  \end{subfigure}

  \vspace{0.2 cm} 
  
  \begin{subfigure}[b]{0.48\linewidth}
    \centering
    \includegraphics[width=\linewidth]{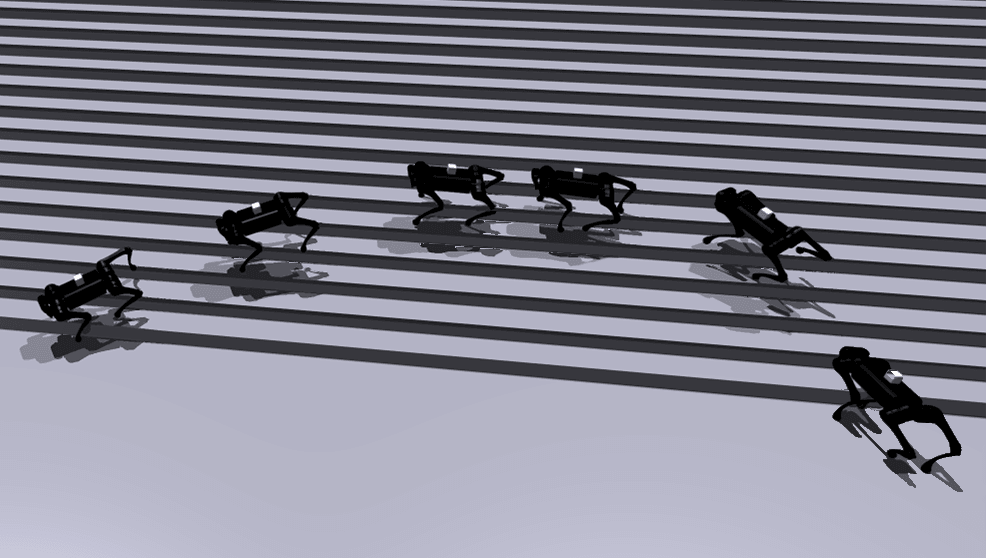}
    \caption{The drifting behavior of our policy on the stairs at 58 °C} 
    \label{fig:fig7b}
  \end{subfigure}
  \hfill
  \begin{subfigure}[b]{0.48\linewidth}
    \centering
    \includegraphics[width=\linewidth]{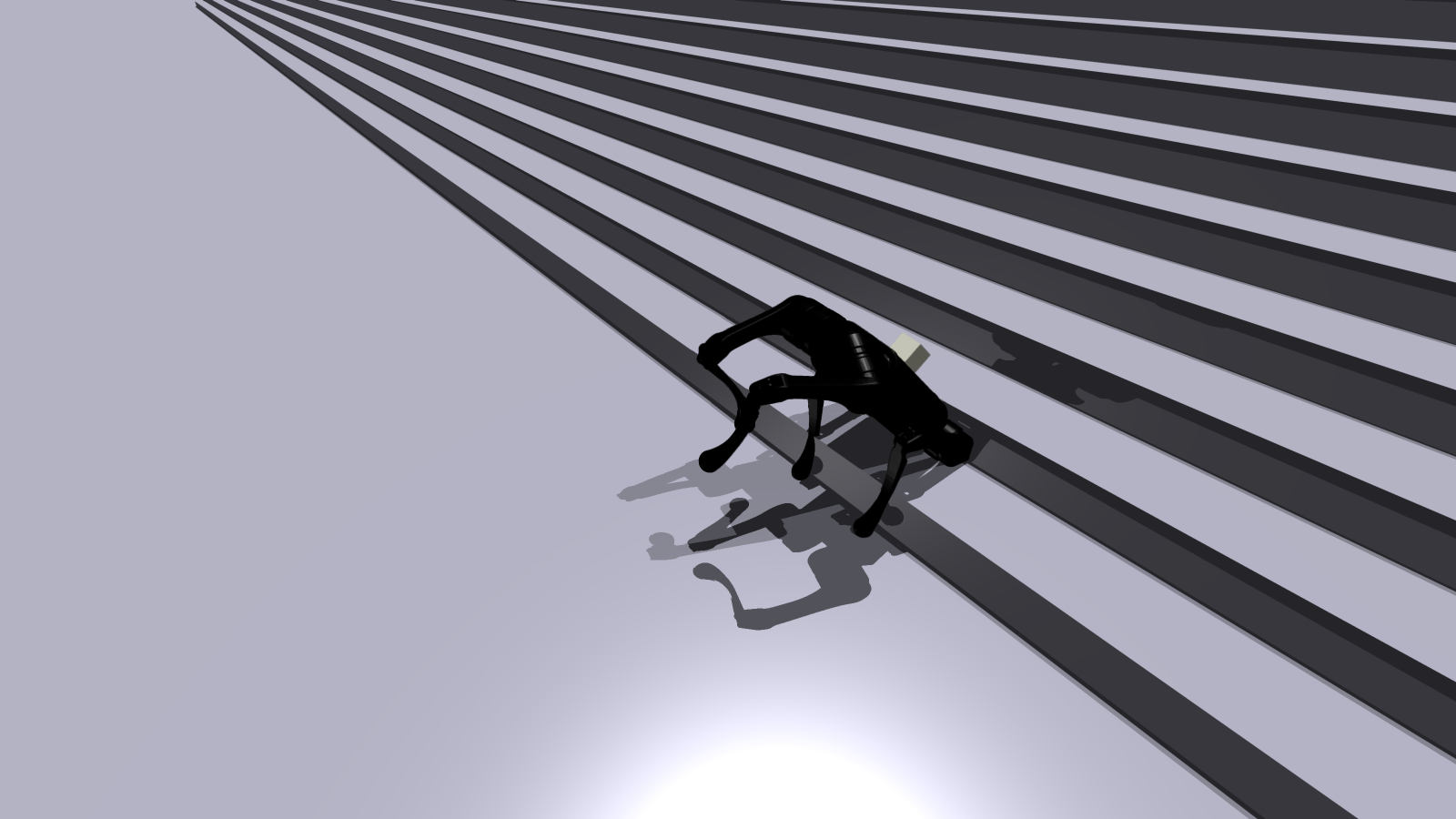}
    \caption{The stuck behavior of MTP on the stairs at 58 °C} 
    \label{fig:fig7c}
  \end{subfigure}

  \caption{Experimental results of the robot carrying a 3 kg payload and following a 1 m/s velocity command on stairs and slopes, with initial motor temperatures of 30 °C, 50 °C, and 58 °C.}
  \label{fig:figure7}
\end{figure}

As shown in Fig.~\ref{fig:fig7a}, NLP exhibited the strongest terrain traversal capability, maintaining a success rate above 99\% at low and medium motor temperatures. However, due to the lack of thermal management, the overheat rate reached approximately 70\% when the initial motor temperature was $58^\circ\text{C}$. MTP consistently resulted in failed and stuck outcomes on stairs, while on slopes, the proportion of drifting reached 50\%, indicating that MTP could not handle these terrains effectively. Although it kept the average peak motor temperature below $60^\circ\text{C}$ at high initial motor temperatures, this came at the cost of losing most of its terrain adaptability.

In contrast, our policy demonstrated terrain traversal performance nearly identical to NLP at low temperatures, with traversal times shown in Table  \ref{tab:table2} comparable to those of NLP. As the temperature increased, the success rate slightly decreased and traversal time extended to around 7.1 s at medium motor temperatures, reflecting the gradual intervention of the residual policy. At high temperatures, our strategy effectively reduced the overheat rate to below 10\% compared to 70\% of the MTP, and kept the average maximum temperature during locomotion around $59^\circ\text{C}$. The most notable phenomenon was the sharp increase in drifting show in Fig.~\ref{fig:fig7b}, accounting for approximately 70\% on stairs and 45\% on slopes. Unlike MTP’s blind stalling show in Fig.~\ref{fig:fig7c}, our strategy allowed deviations in yaw velocity when detecting a rapid temperature rise that could exceed the threshold, to change heading and reducing the continuous rise of motor temperature. This validates that our approach can maintain terrain traversal capability at safe temperatures while ensuring thermal safety.

\begin{table}[t]
\centering
\captionsetup{
    labelsep=newline, 
    justification=centering, 
    singlelinecheck=false,
    font=small,              
    labelfont=normalfont,    
    textfont=sc              
  }
\caption{TERRAIN TRAVERSAL RESULTS}
\label{tab:table2}
\begin{tabular}{>{\centering\arraybackslash}m{0.8cm} >{\centering\arraybackslash}m{0.8cm} >{\centering\arraybackslash}m{1.5cm} >{\centering\arraybackslash}m{1.5cm} >{\centering\arraybackslash}m{1.8cm}}
\toprule
Terrain & Policy & Temperature (High) & Traversal Time (Low) & Traversal Time (Medium) \\
\midrule
\multirow{3}{*}{Stairs} 
 & Ours & $59.256_{-1.026}^{+2.202}$ & $6.704_{-0.764}^{+3.296}$ & $7.094_{-0.754}^{+2.966}$ \\
 & NLP  & $60.727_{-1.716}^{+2.890}$ & $6.702_{-0.682}^{+1.838}$ & $6.706_{-0.746}^{+2.114}$ \\
 & MTP  & $59.228_{-0.991}^{+1.401}$ & -- & -- \\
\midrule
\multirow{3}{*}{Slope} 
 & Ours & $59.209_{-0.828}^{+1.457}$ & $6.630_{-0.670}^{+1.850}$ & $7.108_{-0.768}^{+12.032}$ \\
 & NLP  & $60.689_{-1.657}^{+3.045}$ & $6.727_{-0.547}^{+1.033}$ & $6.724_{-0.564}^{+1.036}$ \\
 & MTP  & $58.458_{-0.353}^{+2.413}$ & -- & -- \\
\bottomrule
\end{tabular}
\end{table}

\begin{figure*}[t]
  \centering
  \begin{subfigure}[b]{0.95\textwidth}
    \centering
    \includegraphics[width=\linewidth]{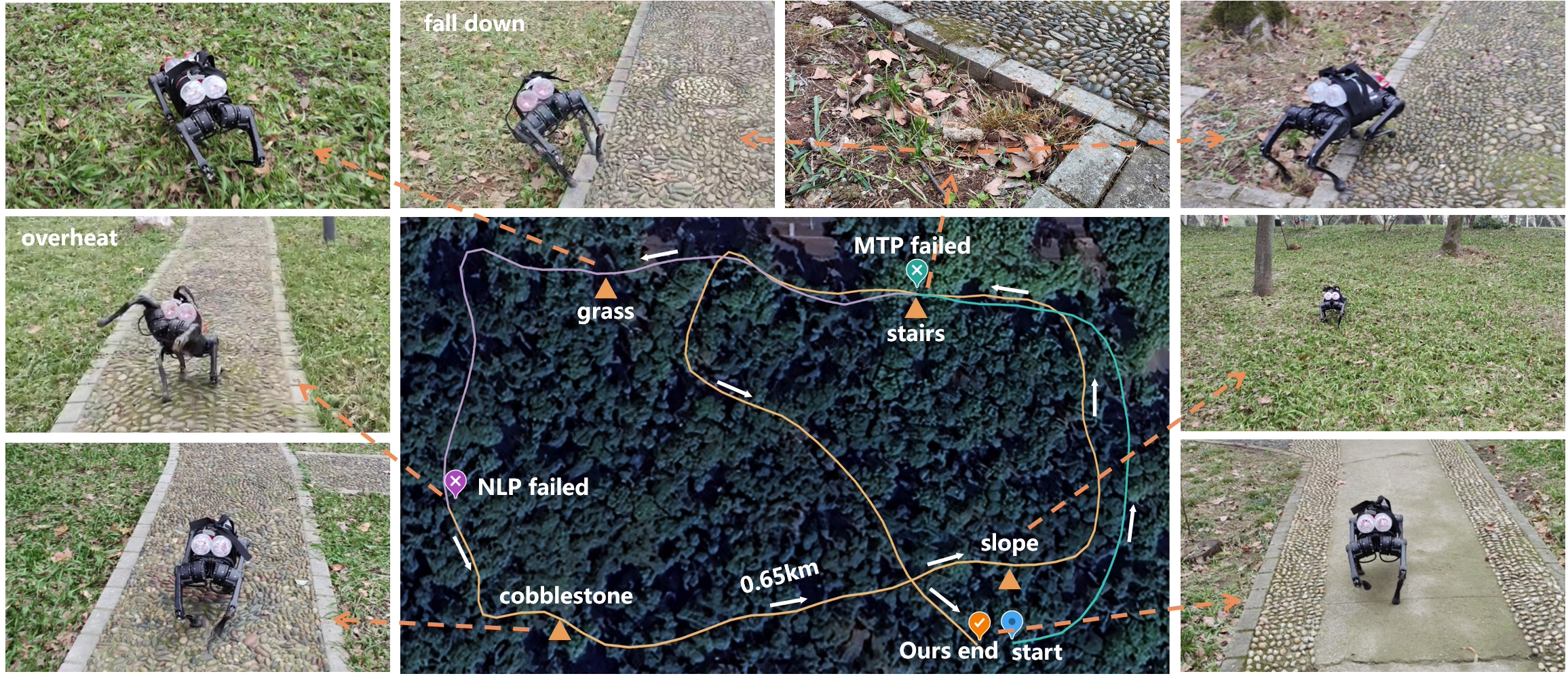}
    \caption{Performance of the three policies along the path}
    \label{fig:fig8a}
  \end{subfigure}

  \vspace{2mm} 

  \begin{subfigure}[b]{0.48\linewidth}
    \centering
    \includegraphics[width=\linewidth]{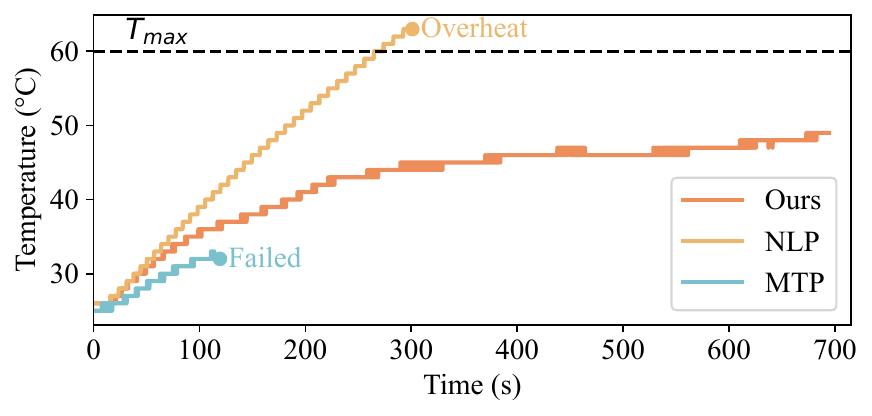}
    \caption{Evolution of the highest motor temperature during locomotion under the three policies}
    \label{fig:fig8b}
  \end{subfigure}
  \hfill
  \begin{subfigure}[b]{0.48\linewidth}
    \centering
    \includegraphics[width=\linewidth]{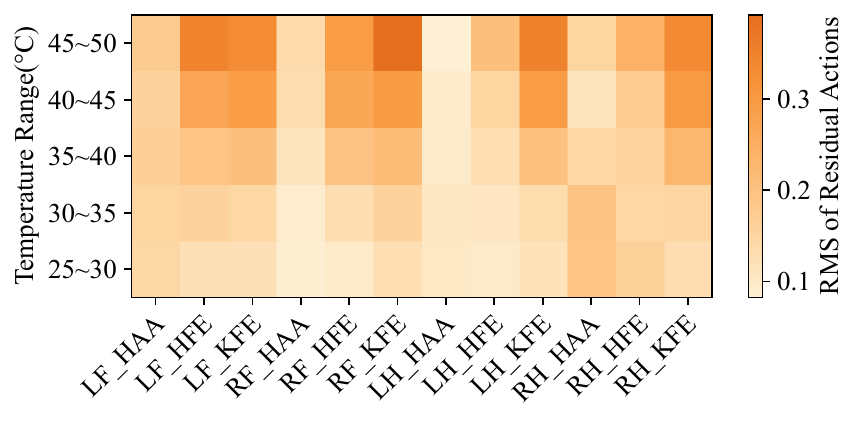}
    \caption{Root mean square (RMS) of the residual actions across different ranges of the highest motor temperature during the experiments; darker colors indicate stronger intervention by the residual policy}
    \label{fig:fig8c}
  \end{subfigure}

  \caption{We conducted a 0.65 km outdoor walking experiment with the Unitree A1 robot carrying a 3 kg payload, over terrain including slopes, grass, cobblestones, and stairs.}
  \label{fig:combined}
\end{figure*}

\subsection{Real-World Outdoor Experiments}

To further evaluate the effectiveness of our proposed method, the Unitree A1 robot carried a 3 kg payload and walked along the path illustrated in Fig.~\ref{fig:fig8a}, comparing the performance of three policies. The outdoor environment included slopes, grass, cobblestones, and stairs, with a total distance of approximately 650 m. The results show that, although NLP could stably traverse those terrains, its lack of thermal management caused the LF\_KFE motor to exceed the threshold temperature in about 5 minutes, according to the temperature measurements. Consequently, the motor could no longer provide the necessary torque for locomotion, leading to task failure after covering less than half of the planned route. In contrast, the robot walked with a high body posture even at low temperatures under MTP (see supplementary video), resulting in a slow temperature rise. However, it failed when going down stairs due to loss of body balance. The robot under our policy successfully traversed all terrains, completing the path in approximately 13 minutes, while keeping the peak motor temperature below $50^\circ\text{C}$. As shown in Fig.~\ref{fig:fig8c}, as the temperature gradually increased, the residual actions progressively grew, providing adaptive thermal regulation demonstrating progressive intervention by the residual policy across all joints, with particularly pronounced adjustments at each leg’s HFE and KFE joints. This enables an effective balance between high-performance locomotion and thermal safety.

\section{CONCLUSIONS}

In this work, we propose a two-stage training framework that incorporates a whole-body thermal model of the quadruped robot. The framework consists of nominal policy pre-training as a locomotion baseline and residual policy learning for thermal management, addressing the challenge of balancing thermal safety and high-performance locomotion. The evaluation results demonstrate the effectiveness of our approach: the robot maintains agility and terrain adaptability at low temperatures while preventing motor overheating at high temperatures. We further validated our approach through hardware experiments on the Unitree A1 quadruped robot, demonstrating its capability for sustained and stable locomotion across diverse terrains. Future research can extend thermal management to the planning layer, generating appropriate commands to the locomotion controller based on the robot’s thermal state. 

\addtolength{\textheight}{-7cm}   

\bibliographystyle{IEEEtran} 
\bibliography{ref}           

\end{document}